\documentclass[conference]{IEEEtran}
\IEEEoverridecommandlockouts
\usepackage{cite}
\usepackage{amsmath,amssymb,amsfonts}
\usepackage{algorithmic}
\usepackage{graphicx}
\usepackage{textcomp}
\usepackage{xcolor}
\usepackage{float}
\usepackage{caption}
\usepackage{subcaption}

\def\BibTeX{{\rm B\kern-.05em{\sc i\kern-.025em b}\kern-.08em
    T\kern-.1667em\lower.7ex\hbox{E}\kern-.125emX}}
\begin{document}

\title{ MCFFA-Net: Multi-Contextual Feature Fusion and Attention Guided Network for Apple Foliar Disease Classification
}

\author{\IEEEauthorblockN{1\textsuperscript{st} Md. Rayhan Ahmed }
\IEEEauthorblockA{\textit{Dept. of Computer Science and Engineering} \\
United International University, Bangladesh \\
rayhan@cse.uiu.ac.bd}
\and
\IEEEauthorblockN{2\textsuperscript{nd} Adnan Ferdous Ashrafi}
\IEEEauthorblockA{\textit{Dept. of Computer Science and Engineering} \\
Stamford University Bangladesh \\
adnan@stamforduniversity.edu.bd}
\and
\IEEEauthorblockN{3\textsuperscript{rd} Raihan Uddin Ahmed}
\IEEEauthorblockA{\textit{Dept. of Electrical and Electronics Engineering} \\
Stamford University Bangladesh \\
raihanahmed95@stamforduniversity.edu.bd}
\and
\IEEEauthorblockN{4\textsuperscript{th} Tanveer Ahmed}
\IEEEauthorblockA{\textit{Dept. of Computer Science and Engineering} \\
Stamford University Bangladesh \\
tanveer.sub.094419@stamforduniversity.edu.bd}
}

\maketitle

\begin{abstract}

Numerous diseases cause severe economic loss in the apple production-based industry. Early disease identification in apple leaves can help to stop the spread of infections and provide better productivity. Therefore, it is crucial to study the identification and classification of different apple foliar diseases. Various traditional machine learning and deep learning methods have addressed and investigated this issue. However, it is still challenging to classify these diseases because of their complex background, variation in the diseased spot in the images, and the presence of several symptoms of multiple diseases on the same leaf. This paper proposes a novel transfer learning-based stacked ensemble architecture named MCFFA-Net, which is composed of three pre-trained architectures named MobileNetV2, DenseNet201, and InceptionResNetV2 as backbone networks. We also propose a novel multi-scale dilated residual convolution module to capture multi-scale contextual information with several dilated receptive fields from the extracted features. Channel-based attention mechanism is provided through squeeze and excitation networks to make the MCFFA-Net focused on the relevant information in the multi-receptive fields. The proposed MCFFA-Net achieves a classification accuracy of 90.86\%.
\end{abstract}


\section{Introduction}
One concerning factor about harvesting apples is that, with time, various diseases have affected production severely. There are many known types of diseases of apple, including viral, viroid, bacterial, fungal, etc. So, it has particularly become necessary to identify these potential diseases beforehand of harvesting and thus ensure a proper treatment plan is executed for apple orchards. The most common diseases that can be identified in an apple tree by examining the leaves of the tree are as follows: mosaic, rust, glomerella leaf spot, black rot, scab, apple litura moth, and apple leaf mites. Out of these, scab due to fungal infection and rust has been of the most concern. The economic loss due to the apple scab alone succumbed to an astounding \$29 million US dollars yearly~\cite{empirical_kwon}. On the other hand, rust in apple can cause severe epidemic outbreaks in an orchard, and thus affected plants needs to be uprooted entirely~\cite{AGRIOS2005265}. Thus, a growing concern about detecting and identifying apple leaf diseases is on the rise.

In recent years, in the agriculture sector, a number of methods for classifying various disease types in crops have been developed as a result of the advancement of various machine learning and computer vision techniques. However, as opposed to typical machine learning algorithms, which require extensive image pre-processing and feature extraction process, the convolutional neural networks (CNNs) can directly learn effective high-level feature representations of different apple foliar diseases from images and model them using their exceptional computational capacity ~\cite{ahmed2021leveraging}.

Using the rising development of deep learning based techniques, the identification and classification of plant leaf diseases are becoming easier and more automated. Various research works are being conducted on image datasets of affected apple leaves. Most recent and noteworthy works feature the prowess of pre-trained deep learning models such as ResNeXt, SE-Net, EfficientNet, VGG, and MobileNet architectures and divergent methods like R-CNN's, transformers, and attention-based mechanisms.

Although several deep learning-based frameworks have been proposed in the literature regarding apple foliar disease classification, various shortcomings and research gaps still remain. Most of the existing studies use only pre-trained architectures as the backbone, directly followed by dense layers for classification. Although pre-trained architectures provide quick convergence of the model and decent accuracy, those models were mostly trained with the ImageNet database, which is very dissimilar to the apple foliar disease images. Besides, features produced from the scab and rust diseased portion of the image are very intricate in detail. Hence, direct implementation of models with only pre-trained architectures struggles to deal with the complex backgrounds, indistinct target boundaries, and diverse and blurry shapes of the region of interest in an image. Low-level shallow features are not sufficient to classify the target object, and the impact of high-level deep features with larger receptive fields proved to be not very significant~\cite{chai2022apple}. That's why it is imperative to extract features from diverse receptive fields with an increasing field of view representation of features for more accurate classification results. In addition, it is also crucial to capture the global dependencies and inter-relationships amongst these receptive fields of diverse sizes through the channel attention mechanism.

In this work, a new transfer learning-based stacked ensemble method named MCFFA-Net has been proposed to classify healthy, and two types of apple foliar disease classes using the Plant Pathology 2020 dataset~\cite{FGVC7}. A novel multi-contextual feature fusion technique along with a channel attention mechanism has been deployed to identify the aforementioned classes. Experimental results after cross-validation have shown promising average accuracy, precision, recall, and F1 score of 90.13\%, 91.31\%, 88.16\%, and 89.67\%, respectively.

The rest of the paper is organized in different sections as follows. Section 2 provides a discussion of the existing literature on apple foliar disease classification. Section 3 provides an overview of the proposed MCFFA-Net model and its incorporated modules. In section 4, we present the experimental result analysis and conclude the paper in section 5.
\vspace{-1.5em}
\section{Related Works}
Deep Learning is a subset of Machine Learning that has gained popularity in significant proportion in the plant pathology domain in recent years and is now being extensively utilized for plant disease classification and diagnosis. Due to the challenge of the scarcity of sufficient datasets, various pre-trained transfer learning-based methods have proved to be an effective tools for the task. Zhong and Zhao~\cite{zhong2020research} employed DenseNet121 architecture as a pre-trained backbone to detect multiple disease types in apple leaves. The authors used 2462 images of six apple leaf diseases and achieved an accuracy of 93.10\%. However, the results are based on a limited validation set consisting of only three diseases grouped into five categories based on disease severity. Khan et al.~\cite{khan2022deep} proposed a two-stage methodology for the classification and detection of diseased apple leaves and achieved an accuracy of 88.50\%. For the initial classification stage, the authors utilize the Xception architecture as the backbone and a global average pooling layer followed by a fully connected layer, and for the detection stage, YOLOv4 was used. In another study, Subetha et al.~\cite{subetha2021comparative} performed a comparative analysis of the performance of ResNet50 and VGG19 algorithms, where both algorithms achieved an accuracy of 87.7\%. Li et al.~\cite{li2021apple} compared the performance of ShuffleNet, MobileNetV3, Vision Transformer, and EfficientNetB0 models for the classification of apple foliar diseases. The authors also studied the impact of various optimizers such as SGD, Adam, RAdam, and Ranger. Storey et al.~\cite{storey2022leaf} utilize instance segmentation to detect leaf and rust disease in apple leave using the Mask R-CNN method. The authors train and assess three different Mask R-CNN-based backbones, namely ResNet-50, MobileNetV3-Large, and MobileNetV3-Large-Mobile, for object segmentation and disease detection tasks. Jiang et al.~\cite{jiang2019real} proposed an improved CNN-based model for real-time detection of apple leaf disease. The model is designed by using the GoogLeNet Inception module, single-shot multibox detector, and rainbow concatenation method~\cite{jeong2017enhancement}, and achieved a mean average detection accuracy of 78.80\%. Li et al.~\cite{li2020apple} enhanced the Faster R-CNN by employing the feature pyramid network and implementing a specific region of interest pooling mechanism. The research revealed that the modified model could detect five apple leaf diseases with an average mean accuracy of 82.28\% under natural conditions.

Recently various attention mechanisms have been utilized in the field of apple foliar disease detection. Wang et al.~\cite{wang2021identification} proposed an attention-based architecture named coordination attention efficientNet model. First, the authors integrate attention block into the EfficientNetB4 used capture both spatial and channel-based feature information. Then, application of depth-wise separable convolution, to the coordination attention convolution network ensures reduction of the number of parameters and avoid the problem of gradient vanishing. In another work, Chai et al.~\cite{chai2022apple} proposed a residual  model for apple leaf disease classification based on pyramidal convolution for multi-scale feature extraction and integrated the squeeze and excitation mechanism~\cite{hu2018squeeze} to improve the weighting of the diseased features.

\section{Materials and Methods}
\vspace{-1em}
\begin{figure}[!htbp]
\centerline{\includegraphics[width= 0.45\textwidth]{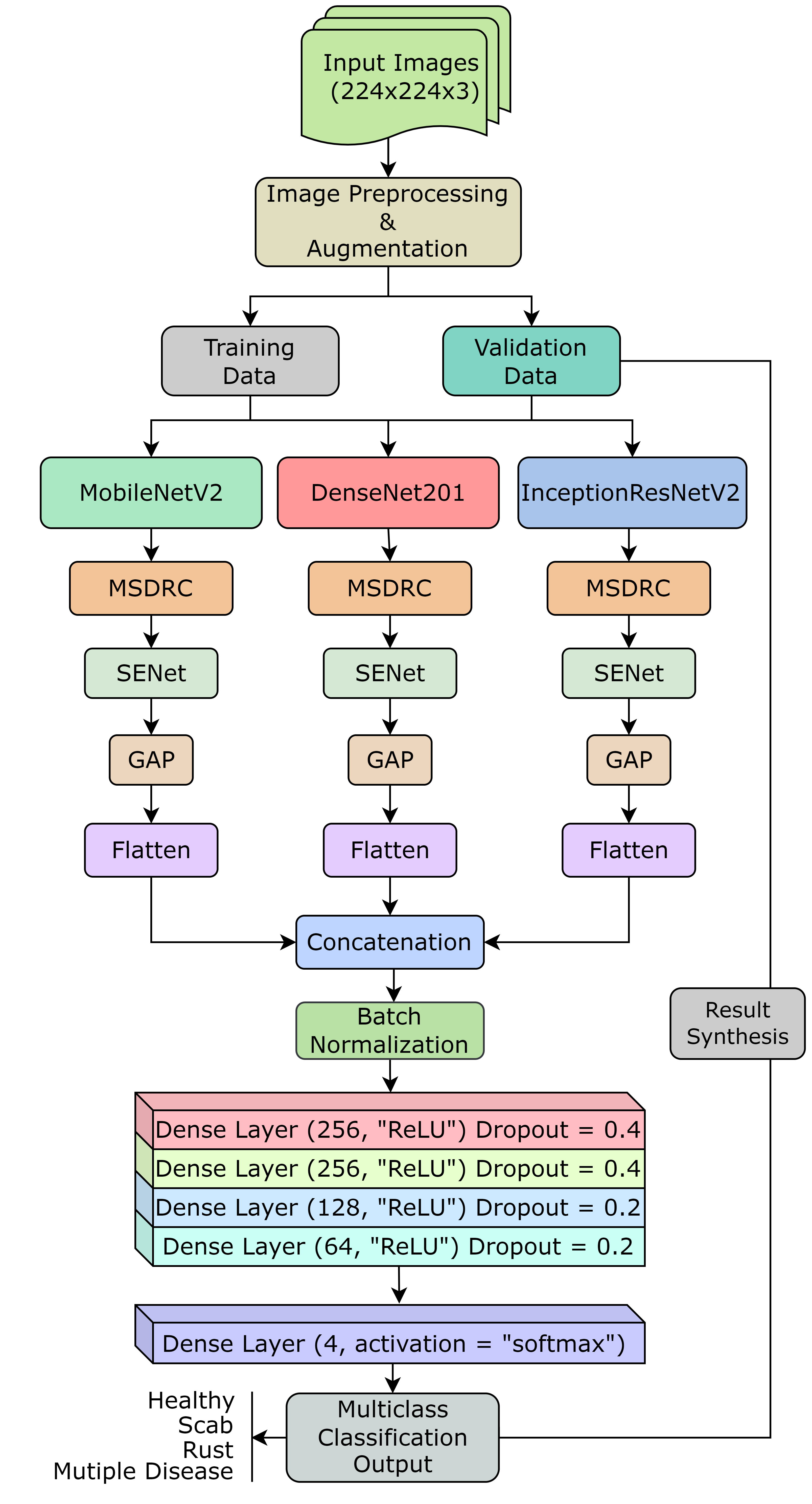}}
\caption{Proposed MCFFA-Net Architecture.}
\vspace{-1em}
\label{fig:proposed_model}
\end{figure}
\begin{figure*}[!htbp]
\centerline{\includegraphics[width=0.85\textwidth]{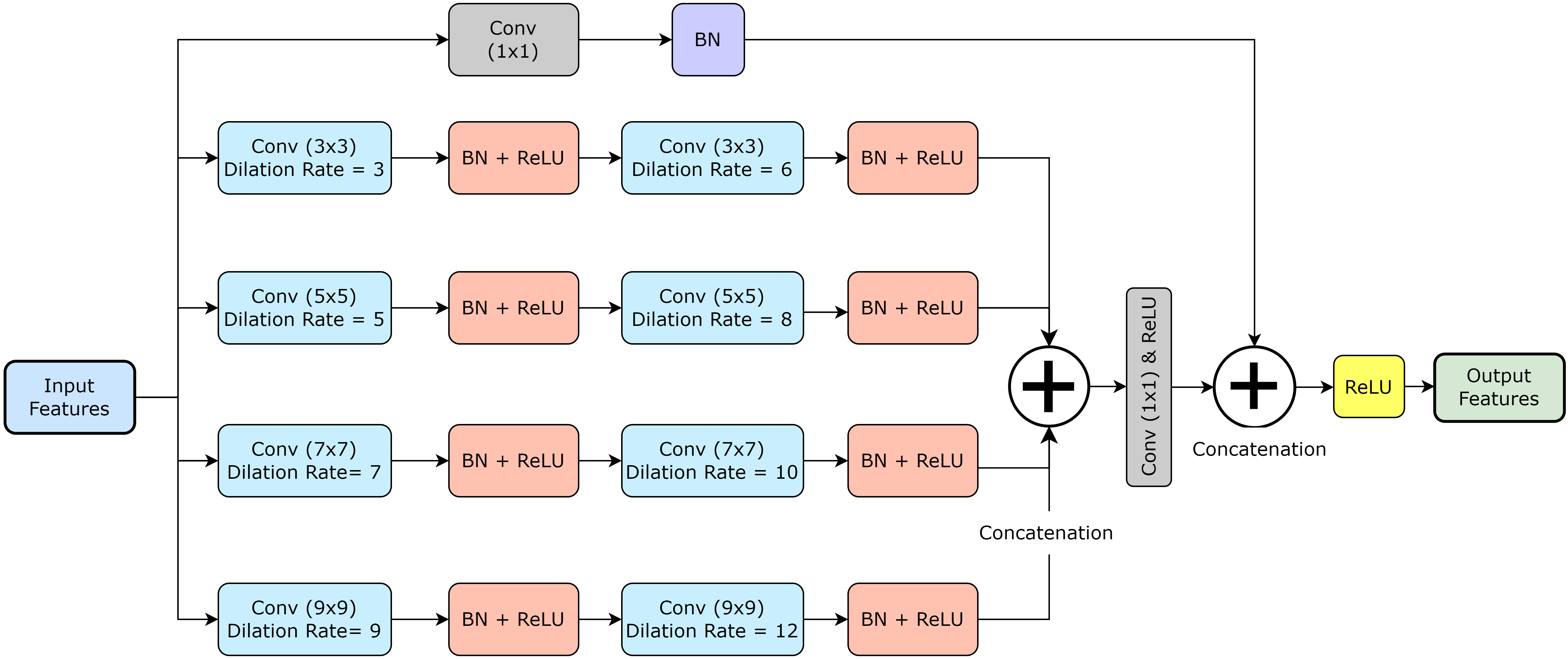}}
\caption{Architecture of Multi-scale dilated residual convolution module.}
\vspace{-1.5em}
\label{fig:msrc}
\end{figure*}
As illustrated in Figure ~\ref{fig:proposed_model}, the proposed MCFFA-Net architecture is based on the transfer learning approach and incorporates three pre-trained architectures, namely MobileNetV2~\cite{sandler2018mobilenetv2}, DenseNet201~\cite{huang2017densely}, and InceptionResNetV2~\cite{szegedy2017inception} as the backbone. All three pre-trained models use the weights of ImageNet and are used to contribute their generalization to the proposed MCFFA-Net architecture. We also employ three multi-scale dilated residual convolutions (MSDRC) modules to capture feature maps of varying sizes and aggregate multi-scale contextual information from each of the aforementioned pre-trained models. After that, the output feature maps of each of the MSDRC modules are passed to the squeeze and excitation network (SE-Net), which provides channel-based attention to the features with higher importance being provided to specific channels over others. Next, we perform a global average pooling (GAP) in all the sub-networks, followed by flattening and concatenation of the features. Then fully connected layers of 256, 256, and 128 neurons with ReLu activation function and dropouts of 0.4, 0.4, and 0.2 are incorporated. The final layer consists of four neurons with each specifying a target class type with the activation of softmax function. The ensemble of these sub-networks is responsible for improved classification accuracy compared to the other state-of-the-art models in the literature.

\subsection{Transfer learning and pre-trained architectures}
Transfer learning is a mechanism for retaining acquired knowledge from one task, $T1$ (i.e., typically pre-trained models), and applying the generated weights from $T1$ to a comparable but distinct task $T2$, with $T1 \neq T2$. It can improve a deep learning-based model's robustness and generalization and assist in learning more effectively and converging quickly. In our proposed MCFFA-Net stacked ensemble architecture, we adopt three pre-trained architectures, namely MobileNetV2, DenseNet201, and InceptionResNetV2, through the transfer learning process. 

MobileNetV2~\cite{sandler2018mobilenetv2} is a lightweight fifty-three-layer deep neural network. Similar to MobileNetV1, MobileNetV2 uses depth-wise separable convolutions. It adds two new features to the existing architecture, the first one is linear bottlenecks across the layers, and the second one is the shortcut connections between the linear bottlenecks. The fundamental concept is that bottlenecks perform the encoding of the model's intermediate inputs and outputs, and similar to traditional residual connections, shortcuts provide faster training and higher accuracy.

Dense Convolutional Network (DenseNet)~\cite{huang2017densely} uses a layer-wise approach to attach each preceding layer to the inbound layers in a feed-forward method. Due to this fashion, DenseNet, not only reduces a substantial number of parameters and vanishing gradient problems but also strengthens feature propagation, mapping, re-usability, and utilization of dense blocks. Having an input size of ($224 \times224 $), the DenseNet201 network learned rich feature representations for a large number of images. Similar to the aforementioned MobileNetV2, it also offers shortcuts to the architecture that can help achieve accuracy with higher values and faster training.

InceptionResNetV2~\cite{szegedy2017inception} is a hybrid Inception style network, which was built as a modification on Inception-V3. This residual inception network uses cheaper Inception blocks than the one used in the original Inception. Each Inception block is followed by a $1 \times 1$ convolution without activation, which scales up the dimensionality. It uses batch normalization on top of traditional layers but does not do so in pooling layers. This network is 164 layers deep and uses a higher number of inception blocks and, consequently higher number of parameters available for training.

\subsection{Multi-scale dilated residual convolution module}

To achieve effective results in image datasets with smaller sizes, it is necessary to extract high-level multi-scale features through different receptive fields. In our proposed architecture, a multi-scale dilated residual convolution (MSDRC) module was applied after each pre-trained model. It helps to reduce saturation and degradation in the learning gradient. The module consists of multiple parallel convolution layers with different kernel sizes of ($3 \times3 $), ($5 \times5 $), ($7 \times7 $), and ($9 \times9 $), respectively. To increase the field of view representation of features and capture more contextual information at the multi-scale, we apply different dilation factors to the convolution operations. Increasing the kernel size in the convolution layers enables the networks to extract a more robust feature representation from multi-scale receptive fields. For the ($3 \times3 $) convolutions, we apply dilation rates of 3 and 6; for ($5 \times5 $) convolutions dilation rates of 5 and 8; for ($7 \times7 $) convolution dilation rates of 7 and 10, and for ($9 \times9 $) convolutions, we apply dilation rates of 9 and 12, respectively. After that, all four feature maps are concatenated together, which leaves us with information on every relevant receptive field, followed by a ($1 \times1 $) convolution and ReLu. We also add a residual shortcut connection, aiming to resolve the vanishing gradient issue of deeper networks~\cite{he2016identity}. It connects the input features and output of the previous ($1 \times1 $) convolution layer. The architecture of the MSDRC module can is demonstrated in Figure \ref{fig:msrc}.

\subsection{Squeeze and excitation-based channel attention}
\begin{figure}[!htbp]
\centerline{\includegraphics[height=4 cm, width=0.50\textwidth]{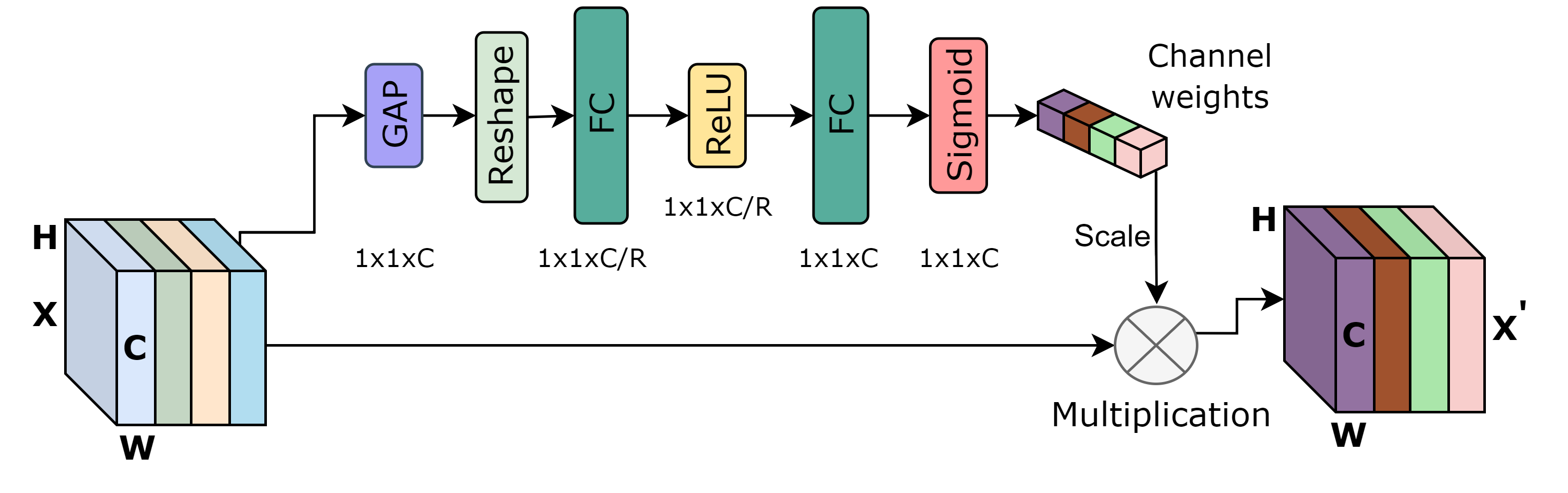}}
\caption{Architecture of the SE-Net module~\cite{hu2018squeeze}.}
\vspace{-1.5em}
\label{fig:senet}
\end{figure}

Squeeze and excitation network (SE-Net)~\cite{hu2018squeeze}, as demonstrated in Figure~\ref{fig:senet} performs dynamic channel-wise feature re-calibration and improves the network’s representative power, and enhances the channel inter-dependencies by providing access to global information. Operations of a SE-Net module are as follows: In order to extract global information from each of the channels of an image, the 2D feature map of each channel of the input feature map ($H \times W$) is first compressed into a real number via the global average pooling method (i.e., squeeze). If channel $u_{i}\subseteq \mathbb{R}^{H \times W}$, then the squeeze element, $z_k$ can be defined using equation \ref{eq:squeeze}.
\begin{equation}
    Z_k = \frac{1}{H \times W}\sum_{i}^{H}\sum_{j}^{W}\textbf{u}_{k}(i,j)
    \label{eq:squeeze}
\end{equation}
To obtain a global statistic for every channel, the ($ H \times W \times C $) image is reduced to the ($ 1 \times 1 \times C $) format. Then, in order to determine the weight of each feature channel, two fully connected layers with ReLU and sigmoid activation functions produce the ‘excitation’ operation. The output is the same number of weights as the input features. The first fully connected layer is utilized for dimensionality reduction by a ratio $R$, and the second fully connected layer is a dimensionality-increasing layer returning to the channel dimension of the previous layer. If $F_1$ and $F_2$ are fully connected layers and sigmoid is $\sigma(.)$ and ReLU operator is $\delta(.)$, then the excitation function can be represented as equation \ref{eq:excite}:
\begin{equation}
    S_{c} = \sigma(F_2\delta(F_{1}Z_{k}))
    \label{eq:excite}
\end{equation}

Lastly, the above-obtained normalized weights are then applied to the features of each channel with the objective of extracting specific information as defined in equation \ref{eq:scale}.
\begin{equation}
    \hat{\textbf{U}} = S_{c} . u_{k}
    \label{eq:scale}
\end{equation}
\vspace{-1.5em}

\section{Experiments and Analysis}

\subsection{Dataset}
\begin{figure}[!htb]
\centerline{\includegraphics[width=0.45\textwidth]{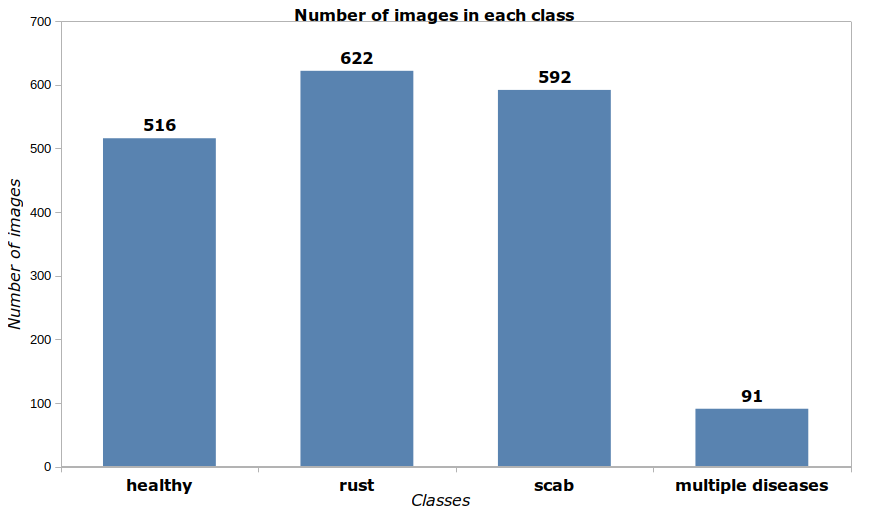}}
\caption{Frequency of images in each class of dataset.}
\vspace{-1.5em}
\label{fig:class_count}
\end{figure}

The dataset employed in this paper is part of the Kaggle Plant Pathology 2020-FGVC7 classification competition~\cite{FGVC7}. A total of 1821 images belonging to four classes: healthy, rust, scab, and multiple diseases, have been resized to $224 \times 224$ pixel to be provided as input to the proposed model. The dataset was divided into 80:20 ratio with 80\% for training (1458 images) and 20\% for validation (363 images). The number of images belonging to each class is provided in the following bar plot of Figure~\ref{fig:class_count}. Due to limited training images, real-time data augmentation was performed using the Keras ImageDataGenerator module, with augmentation techniques like share and zoom range, height and width shift, horizontal and vertical flip, changing the brightness, and random rotation. 

\begin{figure*}[!htb]
     \centering
     \begin{subfigure}[b]{0.45\textwidth}
         \centering
         \includegraphics[width=\textwidth]{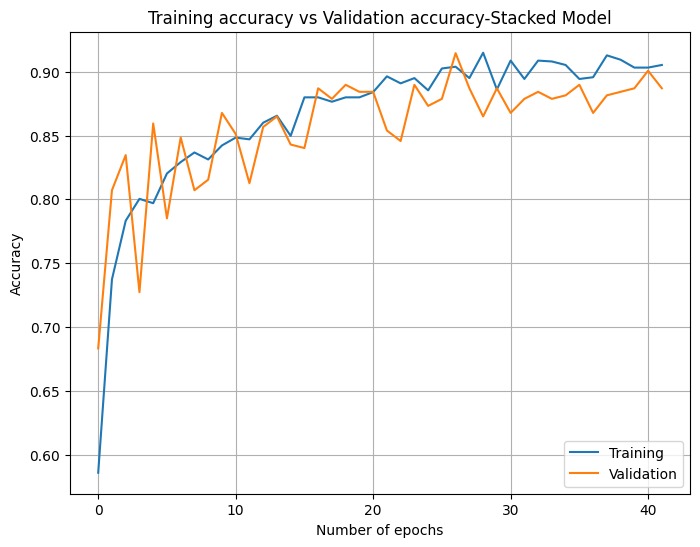}
         \caption{$Validation Accuracy =89.26$}
         \label{fig:acc1}
     \end{subfigure}
     \hfill
     \begin{subfigure}[b]{0.45\textwidth}
         \centering
         \includegraphics[width=\textwidth]{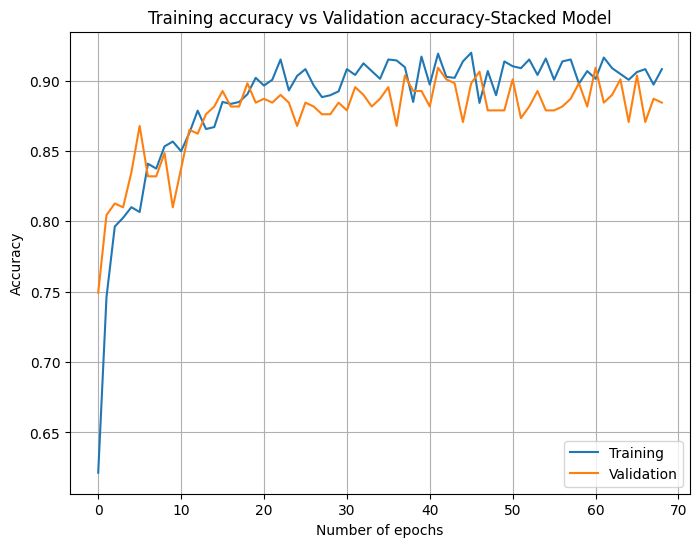}
         \caption{$Validation Accuracy = 90.40$}
         \label{fig:acc2}
     \end{subfigure}
     \vspace{0.1cm}
     \begin{subfigure}[b]{0.45\textwidth}
         \centering
         \includegraphics[width=\textwidth]{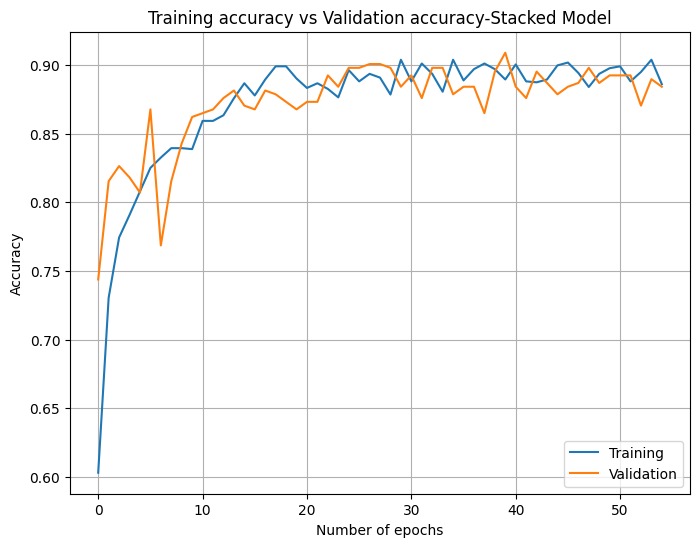}
         \caption{$Validation Accuracy = 89.98$}
         \label{fig:acc3}
     \end{subfigure}
     \hfill
     \begin{subfigure}[b]{0.45\textwidth}
         \centering
         \includegraphics[width=\textwidth]{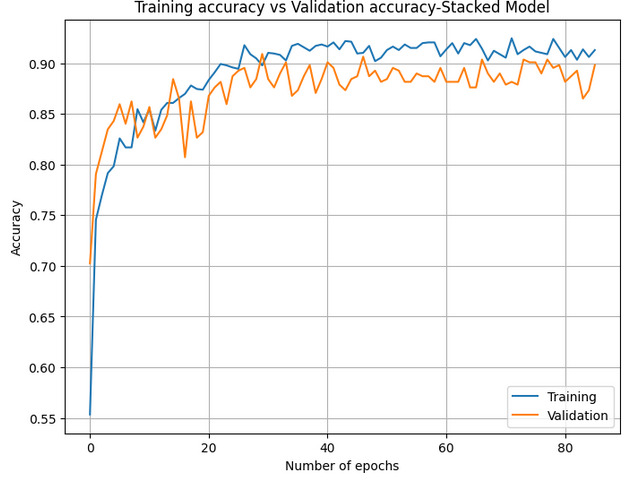}
         \caption{$Validation Accuracy = 90.86$}
         \label{fig:acc4}
     \end{subfigure}
        \caption{Training versus Validation Accuracy graphs in different folds where, (a) Fold 1, (b) Fold 2, (c) Fold 3, and (d) Fold 4 }
        \vspace{-1.5em}
        \label{fig:acc_graphs}
\end{figure*}

\subsection{Optimization and hyper-parameter tuning}
In this research work, a variety of hyper-parameter tuning was done to make sure that the model converges toward the required goal. A learning rate scheduler was deployed, which reduced the learning rate after every $10$ epoch. Next, a reduced learning rate on a plateau was also employed, which monitored the metric val\_loss and reduced the learning rate by a factor of $0.1$ upon staying on a plateau for more than $5$ epochs. For all experimentation, the batch size was fixed at $16$. The initial learning rate was started at $0.001$ and was capped at $10^{-7}$. In order to ensure that loss of computation power was minimized, an early stopping mechanism with a patience of 35 was deployed.

\subsection{Evaluation Metrics}
In order to evaluate the results, several metrics were deployed. All of the metrics originated from the confusion matrix and were computed from the predicted and actual annotated labels of the validation dataset as follows:
\begin{equation}
Accuracy=\frac{TP+TN}{TP+FP+TN+FN}\label{acc}
\end{equation}
\begin{equation}
Precision=\frac{TP}{TP+FP}\label{precision}
\end{equation}
\begin{equation}
Recall=\frac{TP}{TP+FN}\label{recall}
\end{equation}
\begin{equation}
F1=\frac{2\times Precision \times Recall}{Precision+Recall}\label{f1}
\end{equation}

\subsection{Result Analysis}

We investigated the performance by creating a stacked ensemble with various pre-trained architectures as the network backbone along with MSDRC and channel attention modules and dense layers. The values from different metrics as obtained from various experimental setups are specified in Table \ref{tab:result}. Here, the stacked ensemble of MobileNetV2, DenseNet201 and InceptionResNetV2 reached the highest validation accuracy and precision of 90.86\% and 92.07\%, respectively on the validation images. Although, the highest recall and F1 score values, 88.86\% and 90.30\%, respectively, were obtained from one of the other investigated stacked ensemble of MobileNetV2, DenseNet201 and ResNet101V2 pre-trained architectures. In  Table \ref{tab:com} we have performed a comparative analysis of our proposed method with existing architectures which incorporated the Plant Pathology 2020-FGVC7 dataset.

\begin{table}[H]
\caption{Obtained evaluation metric scores using different variants of the proposed multi-contextual architecture. Here, $A =$ MobileNetV2, $B =$ DenseNet201, $C =$ InceptionResNetV2, $D = $ SeResNeXT, $E = $ DenseNet169, $F = $ ResNet101V2, $G = $ Xception, $H = $ InceptionV3}
\begin{center}
\begin{tabular}{|c|c|c|c|c|}
\hline
Ensemble & Accuracy & Precision & Recall & F1\_Score\\
\hline
$A$, $E$, $G$ & 84.85 & 90.03 & 81.00 & 85.15\\
\hline
$A$, $E$, $H$ & 86.50 & 87.97 & 85.08 & 86.44\\
\hline
$A$, $B$, $F$ & 89.81 & 91.88 & 88.86 & 90.30\\
\hline
$A$, $B$, $D$ & 88.43 & 90.84 & 85.08 & 87.81\\
\hline
$A$, $B$, $E$ & 88.15 & 89.76 & 87.38 & 88.52\\
\hline
$A$, $C$, $E$ & 88.98 & 91.26 & 86.23 & 88.93\\
\hline
$A$, $B$, $C$ & \textbf{90.86} & \textbf{92.07} & \textbf{88.46} & \textbf{90.15}\\
\hline
\end{tabular}
\label{tab:result}
\end{center}
\end{table}
\vspace{-1.5em}
\begin{figure}
    \centering
    \includegraphics[width=0.45\textwidth]{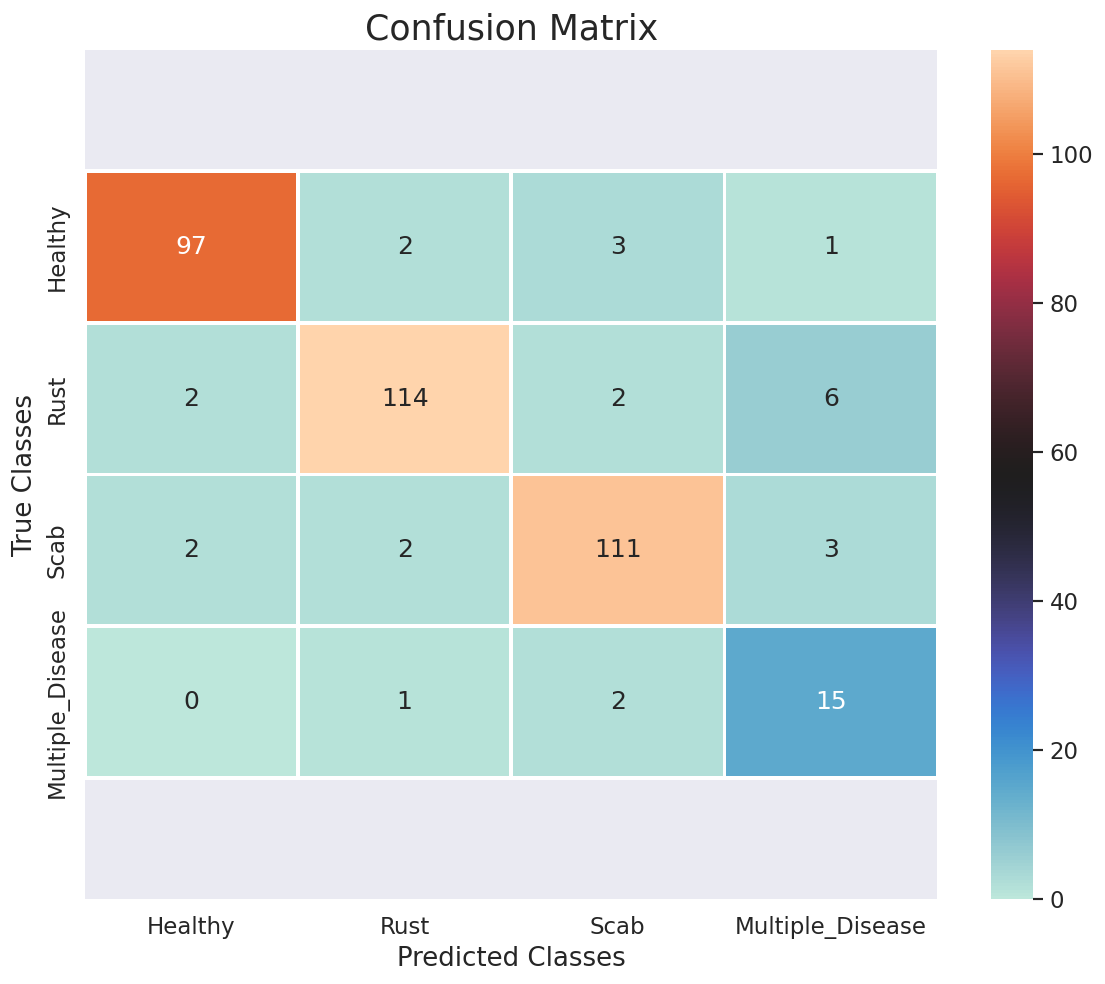}
    \caption{Confusion matrix of MCFFA-Net model in fold-4.}
    \vspace{-1em}
    \label{fig:conf_mat}
\end{figure}

\begin{table}[!htb]
    \caption{Comparison with state-of-the-art methods}
    \centering
    \begin{tabular}{|c|c|}
     \hline
         Model & Accuracy  \\
          \hline
         \cite{khan2022deep} & 88.50 \\
          \hline
          \cite{subetha2021comparative} & 87.70 \\
         \hline
         \cite{jiang2019real} & 78.80 \\      
         \hline
         \cite{li2020apple} & 82.28 \\ 
           \hline
         MCFFA-Net (Ours) & 90.86\\
          \hline
    \end{tabular}
    \vspace{-1em}
    \label{tab:com}
\end{table}
\begin{table}[H]
\caption{Obtained evaluation metric scores in cross-validation study}
\label{tab:k-fold}
\vspace{-1em}
\begin{center}
\begin{tabular}{|c|c|c|c|c|}
\hline
k-Fold & Accuracy & Precision & Recall & F1\_Score\\
\hline
1 & 89.26 & 90.21 & 87.65 & 88.88\\
\hline
2 & 90.40 & 91.14 & 88.34 & 89.72\\
\hline
3 & 89.98 & 91.89 & 88.05 & 89.93\\
\hline
4 & \textbf{90.86} & \textbf{92.07} & \textbf{88.46} & \textbf{90.15}\\
\hline
Avg. & 90.13 & 91.31 & 88.16 & 89.67\\
\hline
\end{tabular}
\label{tab:result2}
\end{center}
\end{table}
\vspace{-1.5em}

\subsubsection*{\textbf{Cross-validation results}}
As per the results in Table \ref{tab:result} we reached the conclusion that the proposed model MCFFA-Net which is the stacked ensemble of MobileNetV2, DenseNet201, and InceptionResNetV2, was the best architecture for the given problem. The experiments were thus repeated for the same architecture four times to ensure the authenticity of the obtained results. In Table \ref{tab:result2}, the metric scores obtained in each fold of the 4-fold cross-validation method are reported. The obtained results indicate that the results obtained from the stacked ensemble MCFFA-Net vary very slightly for different validation subsets. The mean accuracy achieved by the MCFFA-Net is 90.13\%. Almost in each iteration, the model reached an early stopping stage. This ensures that our proposed model was able to converge to such an accuracy within a very limited time. The training versus validation accuracy curve of each fold is presented in Figure~\ref{fig:acc_graphs}.

\section{Conclusion}
In this research work, a robust transfer learning-based stacked ensemble architecture named MCFFA-Net is proposed. In each of the experimental setups, the effectiveness of the proposed model was evident through the general evaluation metrics. A newly designed multi-scale-dilated-residual-convolution module widened the field of view representation of the receptive fields of the channels and provided MCFFA-Net to model features from diverse scales, which is imperative in the plant disease classification tasks. The integration of the SE-Net-based attention module improved the overall performance by capturing relevant information from the multi-scale feature maps. The results obtained from the experiments have concrete indications that MCFFA-Net has very good potential in the classification tasks on apple foliar diseases. Furthermore, in the future, we will work on making the proposed model as lightweight as possible to be integrated into mobile devices.

\bibliographystyle{IEEEtran}
\bibliography{bibliography}

\begin{thebibliography}{10}
\providecommand{\url}[1]{#1}
\csname url@samestyle\endcsname
\providecommand{\newblock}{\relax}
\providecommand{\bibinfo}[2]{#2}
\providecommand{\BIBentrySTDinterwordspacing}{\spaceskip=0pt\relax}
\providecommand{\BIBentryALTinterwordstretchfactor}{4}
\providecommand{\BIBentryALTinterwordspacing}{\spaceskip=\fontdimen2\font plus
\BIBentryALTinterwordstretchfactor\fontdimen3\font minus
  \fontdimen4\font\relax}
\providecommand{\BIBforeignlanguage}[2]{{%
\expandafter\ifx\csname l@#1\endcsname\relax
\typeout{** WARNING: IEEEtran.bst: No hyphenation pattern has been}%
\typeout{** loaded for the language `#1'. Using the pattern for}%
\typeout{** the default language instead.}%
\else
\language=\csname l@#1\endcsname
\fi
#2}}
\providecommand{\BIBdecl}{\relax}
\BIBdecl

\bibitem{empirical_kwon}
D.~Kwon, S.~Kim, Y.~Kim, M.~Son, K.~Kim, D.~An, and B.~Kim, ``An empirical
  assessment of the economic damage caused by apple marssonina blotch and pear
  scab outbreaks in korea,'' \emph{Sustainability}, vol.~7, pp.
  16\,588--16\,598, 12 2015.

\bibitem{AGRIOS2005265}
\BIBentryALTinterwordspacing
G.~N. AGRIOS, ``chapter eight - plant disease epidemiology,'' in \emph{Plant
  Pathology (Fifth Edition)}, 5th~ed., G.~N. AGRIOS, Ed.\hskip 1em plus 0.5em
  minus 0.4em\relax San Diego: Academic Press, 2005, pp. 265--291. [Online].
  Available:
  \url{https://www.sciencedirect.com/science/article/pii/B9780080473789500142}
\BIBentrySTDinterwordspacing

\bibitem{ahmed2021leveraging}
M.~R. Ahmed, ``Leveraging convolutional neural network and transfer learning
  for cotton plant and leaf disease recognition,'' \emph{Int. J. Image. Graph.
  Signal Process}, vol.~13, pp. 47--62, 2021.

\bibitem{chai2022apple}
H.~Chai, Z.~Guo, and J.~Yang, ``Apple leaf disease recognition based on
  attention mechanics and multi-scale feature fusion,'' in \emph{2022 14th
  International Conference on Machine Learning and Computing (ICMLC)}, 2022,
  pp. 368--374.

\bibitem{FGVC7}
\BIBentryALTinterwordspacing
``Plant pathology 2020 - fgvc7 | kaggle.'' [Online]. Available:
  \url{https://www.kaggle.com/competitions/plant-pathology-2020-fgvc7}
\BIBentrySTDinterwordspacing

\bibitem{zhong2020research}
Y.~Zhong and M.~Zhao, ``Research on deep learning in apple leaf disease
  recognition,'' \emph{Computers and Electronics in Agriculture}, vol. 168, p.
  105146, 2020.

\bibitem{khan2022deep}
A.~I. Khan, S.~Quadri, S.~Banday, and J.~L. Shah, ``Deep diagnosis: A real-time
  apple leaf disease detection system based on deep learning,'' \emph{Computers
  and Electronics in Agriculture}, vol. 198, p. 107093, 2022.

\bibitem{subetha2021comparative}
T.~Subetha, R.~Khilar, and M.~S. Christo, ``A comparative analysis on plant
  pathology classification using deep learning architecture--resnet and
  vgg19,'' \emph{Materials Today: Proceedings}, 2021.

\bibitem{li2021apple}
L.~Li, S.~Zhang, and B.~Wang, ``Apple leaf disease identification with a small
  and imbalanced dataset based on lightweight convolutional networks,''
  \emph{Sensors}, vol.~22, no.~1, p. 173, 2021.

\bibitem{storey2022leaf}
G.~Storey, Q.~Meng, and B.~Li, ``Leaf disease segmentation and detection in
  apple orchards for precise smart spraying in sustainable agriculture,''
  \emph{Sustainability}, vol.~14, no.~3, p. 1458, 2022.

\bibitem{jiang2019real}
P.~Jiang, Y.~Chen, B.~Liu, D.~He, and C.~Liang, ``Real-time detection of apple
  leaf diseases using deep learning approach based on improved convolutional
  neural networks,'' \emph{IEEE Access}, vol.~7, pp. 59\,069--59\,080, 2019.

\bibitem{jeong2017enhancement}
J.~Jeong, H.~Park, and N.~Kwak, ``Enhancement of ssd by concatenating feature
  maps for object detection,'' \emph{arXiv preprint arXiv:1705.09587}, 2017.

\bibitem{li2020apple}
X.~Li, S.~Li, and B.~Liu, ``Apple leaf disease detection method based on
  improved faster r\_cnn,'' \emph{Comput. Eng.}, vol.~46, no.~11, pp. 59--64,
  2020.

\bibitem{wang2021identification}
P.~Wang, T.~Niu, Y.~Mao, Z.~Zhang, B.~Liu, and D.~He, ``Identification of apple
  leaf diseases by improved deep convolutional neural networks with an
  attention mechanism,'' \emph{Frontiers in Plant Science}, p. 1997, 2021.

\bibitem{hu2018squeeze}
J.~Hu, L.~Shen, and G.~Sun, ``Squeeze-and-excitation networks,'' in
  \emph{Proceedings of the IEEE conference on computer vision and pattern
  recognition}, 2018, pp. 7132--7141.

\bibitem{sandler2018mobilenetv2}
M.~Sandler, A.~Howard, M.~Zhu, A.~Zhmoginov, and L.-C. Chen, ``Mobilenetv2:
  Inverted residuals and linear bottlenecks,'' in \emph{Proceedings of the IEEE
  conference on computer vision and pattern recognition}, 2018, pp. 4510--4520.

\bibitem{huang2017densely}
G.~Huang, Z.~Liu, L.~Van Der~Maaten, and K.~Q. Weinberger, ``Densely connected
  convolutional networks,'' in \emph{Proceedings of the IEEE conference on
  computer vision and pattern recognition}, 2017, pp. 4700--4708.

\bibitem{szegedy2017inception}
C.~Szegedy, S.~Ioffe, V.~Vanhoucke, and A.~A. Alemi, ``Inception-v4,
  inception-resnet and the impact of residual connections on learning,'' in
  \emph{Thirty-first AAAI conference on artificial intelligence}, 2017.

\bibitem{he2016identity}
K.~He, X.~Zhang, S.~Ren, and J.~Sun, ``Identity mappings in deep residual
  networks,'' in \emph{European conference on computer vision}.\hskip 1em plus
  0.5em minus 0.4em\relax Springer, 2016, pp. 630--645.

\end{thebibliography}

\end{document}